%% file: main.tex
\documentclass[runningheads]{llncs}

\usepackage{eccv}

\usepackage{graphicx}
\usepackage{booktabs}
\usepackage{glossaries}
\usepackage{wrapfig}

\usepackage[accsupp]{axessibility}  

\usepackage[pagebackref,breaklinks,colorlinks]{hyperref}

\usepackage{orcidlink}

\newacronym{tsdf}{TSDF}{Truncated Signed Distance Function}
\newacronym{mlp}{MLP}{Multi-layer Perceptron}
\newacronym{sfm}{SfM}{Structure-from-Motion}
\newacronym{fif}{FIF}{Fisher Information Fields}
\newacronym{gnn}{GNN}{Graph Neural Network}
\newacronym{cnn}{CNN}{Convolutional Neural Network}

\newlength{\originalbelowcaptionskip}
\newlength{\originalabovecaptionskip}
\input{notation.tex}

\begin{document}

\title{Learning Where to Look: Self-supervised Viewpoint Selection for Active Localization using Geometrical Information} 

\titlerunning{Learning Where to Look}

\author{Luca Di Giammarino\inst{1}\orcidlink{0000-0002-1332-3496} \and
Boyang Sun\inst{2}\orcidlink{0009-0009-0166-6465} \and
Giorgio Grisetti\inst{1}\orcidlink{0000-0002-8038-9989} \and
Marc Pollefeys\inst{2,3}\orcidlink{0000-0003-2448-2318} \and
Hermann Blum\inst{2, 4}\orcidlink{0000-0002-1713-7877} \and
Daniel Barath\inst{2}\orcidlink{0000-0002-8736-0222}}

\authorrunning{Di Giammarino et al.}

\institute{$^1$ Sapienza University of Rome \quad $^2$ ETH Z\"{u}rich \quad $^3$ Microsoft \quad $^4$ Uni Bonn}

\maketitle

\begin{abstract}
Accurate localization in diverse environments is a fundamental challenge in computer vision and robotics. The task involves determining a sensor's precise position and orientation, typically a camera, within a given space. Traditional localization methods often rely on passive sensing, which may struggle in scenarios with limited features or dynamic environments. In response, this paper explores the domain of active localization, emphasizing the importance of viewpoint selection to enhance localization accuracy. Our contributions involve using a data-driven approach with a simple architecture designed for real-time operation, a self-supervised data training method, and the capability to consistently integrate our map into a planning framework tailored for real-world robotics applications. Our results demonstrate that our method performs better than the existing one, targeting similar problems and generalizing on synthetic and real data. We also release an open-source implementation to benefit the community at \url{www.github.com/rvp-group/learning-where-to-look}.
\end{abstract}

\begin{figure}
  \centering
  \includegraphics[width=\textwidth]{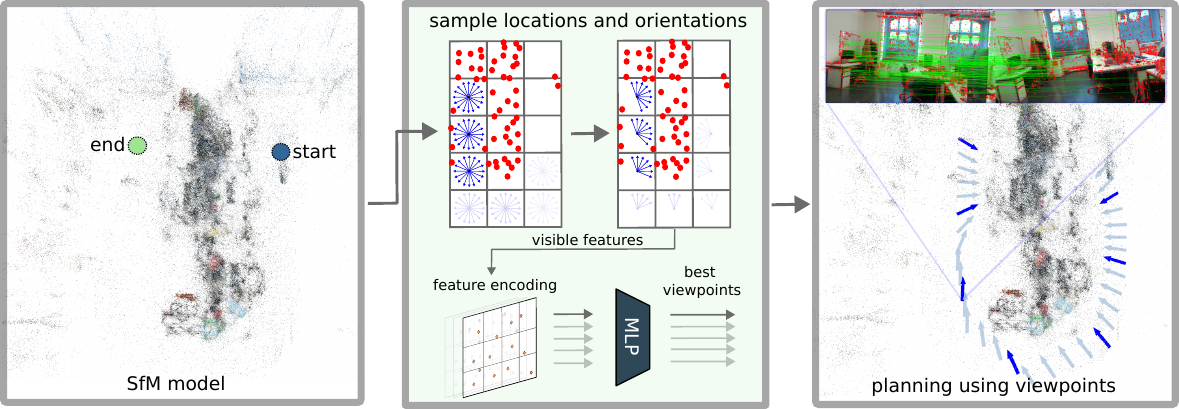}
  \caption{\textbf{Pipeline.} Given a \gls{sfm} model, we aim to learn the camera viewpoint that can be employed to maximize the accuracy in visual localization. Our methodology requires first sampling the camera locations and orientation, calculating the best visibility orientation for each location, and learning active viewpoint through a \gls{mlp} encoder. The illustration above shows our full pipeline predicting active viewpoints for visual localization embedded into a planning framework.}
  \label{fig:motiv}
  \vspace{-0.4cm}
\end{figure}

\section{Introduction}
\label{sec:intro}

Localization and mapping are fundamental building blocks of autonomous systems. 
Localization aims at determining the camera position within an environment, while mapping involves estimating a comprehensive representation of the surrounding space. 
These capabilities enable agents to navigate and operate effectively in unexplored and dynamic settings, with applications ranging from autonomous vehicles \cite{brizi2024vbr}, precision agriculture \cite{saraceni2023agrisort}, and augmented reality \cite{hanlon2023active}.

Decoupling localization and mapping is often advantageous due to the distinct nature of these processes, allowing for separate optimization. 
Offline mapping computation enhances accuracy by overcoming real-time processing constraints and generating a map before the actual operation. 
However, even with offline computation, discrepancies in accuracy may exist across the map, impacting certain areas more than others.
For example, these differences can stem from non-uniformly distributed observations within the space.
In such map regions characterized by heterogeneous accuracy, traditional localization methods may encounter challenges in delivering precise position information \cite{zhang2020fisher}. 
The difficulty lies in discerning meaningful features within the environment to facilitate accurate localization, giving rise to the concept of active localization \cite{burgard1997active}.

Active localization entails purposeful selection and observation of specific environmental features to augment localization accuracy. 
In contrast to passive sensing, where an agent merely observes its surroundings, active localization allows the agent to actively seek out and concentrate on stable features. 
This approach significantly enhances the ability to determine an accurate position, particularly in complex or adverse environments.

This active approach to localization ensures adaptability to diverse surroundings and finds applications in collaborative scenarios. For instance, in human-robot collaboration, active localization may involve seeking input from human operators to identify critical features or areas of interest \cite{hanlon2023active}. 
As technology progresses, active localization plays a pivotal role in enabling consistent robotic operation across a broad spectrum of environments, addressing challenges, and advancing the field of perception and robotics.

The concept of active localization encompasses various interpretations, ranging from enhancing map representation \cite{zhang2020fisher, hanlon2023active} to optimizing localization accuracy through robot planning \cite{roy1999coastal,costante2016perception}. In the context of map representation, studies have employed hand-crafted techniques, leveraging tools like the Fisher information matrix and optimality theory \cite{zhang2020fisher}. 
Conversely, approaches grounded in data-driven methodologies, which showed accurate results, necessitate collecting the training data manually \cite{hanlon2023active}. 
This study addresses both aspects, proposing a map representation that identifies accurate active viewpoints. These viewpoints serve the dual purpose of enhancing localization accuracy and being readily accessible for efficient planning, thus encompassing the limitations observed in existing proposed works.

Inspired by \cite{zhang2020fisher} and \cite{hanlon2023active}, our primary focus is to extract valuable information from the environment's geometry, aiming to enhance active visual localization, enabling the integration of our map representation into a motion planning framework for robotics applications. The contributions of our work include a data-driven approach employing a compact architecture designed for real-time operation, a self-supervised data training method, a map representation facilitating multiple active viewpoints at specific locations in space, and the capability to compactly embed our map into a planning framework tailored for real-world robotics applications. A visual summary of our pipeline is depicted in \figref{fig:motiv}.

\section{Related Works}
\label{sec:related}

Visual localization describes the task of estimating the camera position and orientation for a query RGB/RGB-D image in a known scene (with databases). This task has gathered significant focus, particularly on improving localization accuracy from a certain perspective.  Research within this field generally falls into one of two primary categories: the direct (or one-step) approach and the two-step approach. The first aims to directly estimate the camera pose from the query frame~\cite{sattler2019understanding,ding2019camnet,yan2022crossloc,wang2020atloc,xue2020learning,moreau2022lens,shavit2021learning,shavit2022camera,chen2022dfnet}. This is also known as pose regression. Learning-based models are increasingly being incorporated into this methodology, enhancing robustness and precision by integrating with conventional processes. 
The two-step approach initially identifies correspondences between the query frame and the database, followed by the estimation of the camera pose through optimization. These correspondences can be visual features~\cite{sattler2011fast,sattler2012improving,sattler2016efficient,sattler2017large,dusmanu2019d2,sarlin2019coarse,sarlin2021back,panek2022meshloc}, or dense correspondence between every pixel of the image~\cite{brachmann2018learning,brachmann2019expert,yang2019sanet,cavallari2019real,li2020hierarchical,tang2021learning,dong2022visual}.

All the above approaches are \textit{passive localization}, implying no active decision-making regarding the camera's viewpoint. 
Rather, the focus is on utilizing the captured image for accurate and efficient localization. 
Some studies choose a different perspective to improve the performance of visual perception, such as in the case of visual localization. 
The agent can adjust its sensors autonomously, aiming to enhance perception from alternative viewpoints. This approach is known as active perception, which has been an open research area for over twenty years. Particularly within the domain of mobile autonomous agents, active perception strategies are frequently combined with planning or navigation modules to improve outcomes in tasks like visual localization. This integration enables the agent not only to perceive its environment more effectively but also to make informed decisions about its movements to optimize perception results, \eg~\cite{roy1999coastal,costante2016perception,zhang2018perception}, surface converging~\cite{6631022,fontanelli2009visual,7989531,10161136}. 

In the specific field of active localization, research has been relatively limited. A significant portion of the work in this area focuses on active viewpoint selection through various geometric evaluation metrics. These metrics assess the effectiveness of a selected viewpoint in terms of its impact on visual localization performance. A noteworthy example of such research is~\cite{zhang2020fisher}. It proposes an efficient way of calculating the Fisher information in a 3D environment and uses this quantity to find camera poses that maximize the visibility and vicinity of feature landmarks. More recent works have tried to utilize additional information from visual appearances, such as semantics from the image~\cite{Bartolomei2021Semantic-awareLearning}. Most of the works in this category design a hand-crafted metric that links the geometry and appearance with the performance, such as visual localization accuracy. In contrast, we propose to learn from the distribution of the 3D landmarks and their contribution to the visual localization task.  

Learning-based approaches for active visual localization have recently started to be investigated but mostly rely on end-to-end approaches. PoseNet~\cite{kendall2015posenet}, and one of its extensions \cite{clark2017vidloc} use a convolutional network to implicitly represent the scene, mapping a single monocular image to a 3D pose (position and orientation). \cite{chaplot2018active} proposed a perceptual model to estimate the belief of the current robot state and a policy model over the current belief to localize accurately. The authors in \cite{fang2022towards} propose an uncertainty-driven policy model to plan a camera path for localization. Differently, \cite{lodel2022look} focuses on safety guarantee and proposes a policy model that recommends a collision-free viewpoint while maximizing the information gained from the observation. Works using a reinforcement learning framework are usually tightly coupled with the action execution of the agent, which makes it hard to train and generalize. A recent interesting study to learn active viewpoints shows good results when the problem is cast to classification \cite{hanlon2023active}. However, this work requires user-selected labels, which might be hard and time-consuming to obtain and does not consider an overall representation of the map or involve motion planning.  
   
In this work, we focus on extracting useful information from the environment's geometry to enhance active visual localization with the possibility of embedding our map representation into a motion planning framework for robotics applications.
The contributions of our work are as follows:
\begin{itemize}
\item a data-driven approach relying only on a compact architecture designed for real-time operation in known environments;
\item a way to learn in a self-supervised manner;
\item a map representation that allows for a set of locations in the space to have one or more active viewpoints available;
\item an open-source implementation available at \url{www.github.com/rvp-group/learning-where-to-look}.
\end{itemize}
The proposed method can easily be embedded into a planning framework for real robotics applications.

\begin{figure}[t]
  \centering
  \includegraphics[width=0.99\textwidth]{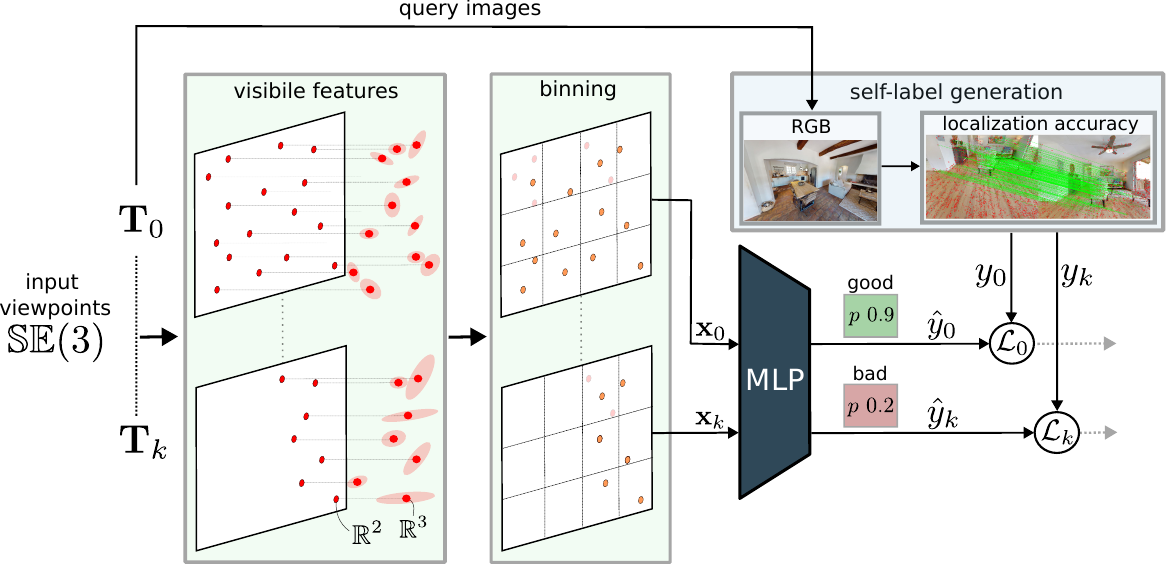}
  \caption{\textbf{Learning active viewpoints.} Given a set of camera poses parameterized as homogeneous transformation matrices, obtained as explained in \secref{sec:sampling} and visibility information (3D landmarks and their projections), our goal is to develop a scoring function that discerns the suitability of a camera viewpoint for visual localization. We first identify visible data from each camera view to achieve this, as elaborated in \secref{sec:visibility}. Subsequently, we encode this visible data through image binning for a fixed input size. The encoded information is then fed into a \gls{mlp} encoder, which predicts the quality of the viewpoint for localization. This learning process, detailed in \secref{sec:nn}, is supervised by consistently providing the camera position, querying an RGB image through a simulator, and directly registering this image against a \gls{sfm} model.}
  \label{fig:learning}
\end{figure}

\section{Learning Where To Look}
\label{sec:our}
We present our viewpoint selection approach in this section. The core of our approach is a learning-based viewpoint evaluation model. The model predicts a visual localization score for an input viewpoint, indicating its efficacy for an accurate localization result. The whole approach follows a ``sampling-and-evaluation'' pipeline. We design a compact workflow to sample candidates' viewpoints from a given 3D environment and construct the input features of the certain viewpoint to pass into the model. A similar data acquisition structure is also applied during training to achieve self-supervision. This section starts with formulating the viewpoint selection task, followed by introducing our major components for building our data collection and learning pipeline, including initial viewpoint sampling, visibility check, and model training.

Our goal is to find a set of camera viewpoints within a sparse set of landmarks or point cloud $\cP = \{ \bl_0, \dots, \bl_L \}$ with each landmark $\bl \in \bbR^3$ that when employed during localization allow to maximize its accuracy. We discretize the orientations 
$\cR = \{\bR_0, \dots, \bR_N\} \in \bbSO(3)^N$ and the locations $\cV = \{\bt_0, \dots, \bt_M\} \in \mathbb{R}^{3\times M}$, using spherical sampling and a voxel grid, respectively.
Using the following discretization, we create the set of camera viewpoints parameterized as homogeneous transformation matrices where an element is represented by $\bT_{ij}$ and is parameterized as follows:  
\begin{equation}
        \bT_{ij} = \begin{bmatrix}
        \bR_i & \bt_j\\
        \bzero & 1 \\
        \end{bmatrix}, \qquad
        \begin{matrix}
         i &= 0, \dots, N \\
         j &= 0, \dots, M    
        \end{matrix}
\end{equation}
Given these two input sets, for each position $\bt_j$ and orientation $\bR_i$, we aim to learn a scoring function that evaluates their suitability for visual localization. Let this scoring function be $\mathbf{f}_{\cP}(\bR_i, \bt_j) \rightarrow [0,1]$. Note that this map representation is fundamental to self-generate data to implicitly learn $\mathbf{f}_{\cP}(\bR_i, \bt_j)$ and can be employed for inference at constant time (\ie~for path planning). A summary of our learning strategy is illustrated in \figref{fig:learning}.

In the next two sections, we present how we sample the 3D locations and orientations composing our map (\secref{sec:space-sampling}).  After, we discuss how the best set of orientations $\cQ$ for each camera location is selected based on landmarks visibility (\secref{sec:visibility}). Finally, in \secref{sec:nn} we present how we learn the function $\mathbf{f}_{\cP}$ to select what viewpoints in the map are more suitable for localization, through a lightweight \gls{mlp}.

\subsection{Sampling}
\label{sec:sampling}
\begin{figure}[t]
  \centering
  \includegraphics[width=0.99\textwidth]{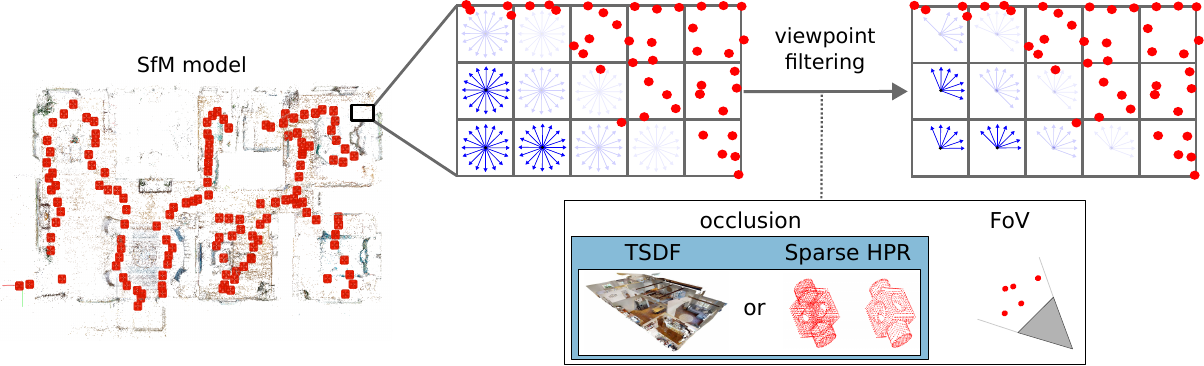}
  \caption{\textbf{Camera viewpoint generation.} We represent our map as a discrete voxel grid $\cV$ and a discrete set of orientations $\cR$ constructed within the boundaries of a 3D reconstruction, \eg, coming from a \gls{sfm} method. 
  We filter the best directions from each camera location in the voxel grid based on visibility $\cQ$, gradually removing occlusions. The illustration is done in 2D for ease of visualization.}
  \label{fig:sampling}
  \vspace{-0.2cm}
\end{figure}

In this section, we detail the process of sampling locations $\cV$ and orientations $\cR$. This enables us to establish a discrete representation of the map, which is essential for localization and applies during both the training and inference phases. During training, this representation becomes necessary for facilitating the self-generation of data. During inference, this representation can be advantageous for planning; however, it is not strictly necessary, given that our learning strategy can effectively utilize any specified position $\mathbf{t}$ and orientation $\mathbf{R}$ of the camera. This procedure is detailed below and illustrated in \figref{fig:sampling}.

\subsubsection{Location.}
\label{sec:space-sampling}
Given as input map a sparse set of landmarks or point cloud $\cP$ and, the 3D boundary of the map, we construct a voxel grid $\cV$ to sample camera positions $\bt$ in the center of each voxel or camera bucket. The resolution of the voxel grid can be arbitrarily determined. In our experiments, we split the 3D space in a $8 \times 8 \times 8$ grid, representing each voxel as a cuboid. 
A mapping between the camera position and their location in memory is stored in a hash table, allowing $O(1)$ insertions and look-ups.

\subsubsection{Orientation.}
\label{sec:orientation-sampling}
\begin{figure}[t]
  \centering
  \begin{subfigure}{0.25\textwidth}
    \includegraphics[width=\linewidth]{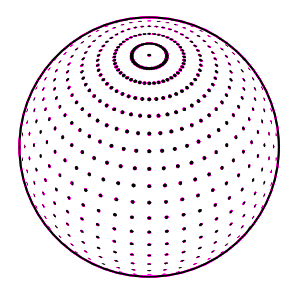}
  \end{subfigure}
  \hspace{2cm}
  \begin{subfigure}{0.25\textwidth}
    \includegraphics[width=\linewidth]{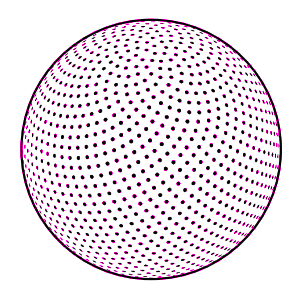}
  \end{subfigure}
  \caption{\textbf{Spherical sampling methods.} The left plot shows classical azimuth-elevation sampling. The right one is Fibonacci sampling. We employed the technique on the right, given the more uniform distributed pattern.}
  \label{fig:fibonacci}
  \vspace{-0.2cm}
\end{figure}

This step is required to generate the set of discrete orientations $\cR$. We only consider azimuth $\theta$ and elevation $\phi$ of $S^2$ from $\bbSO(3)$ to determine each rotation, omitting the rotation around the camera optical-axes. For sampling, we employ the \textit{spherical Fibonacci sampling} algorithm \cite{gonzalez2010measurement}. This sampling is organized in a closely wound generative spiral, where each point is positioned within the largest gap between the preceding points. Different from the classic azimuth–elevation sampling where equal angles of azimuth and elevation separate each point, this sampling exhibit a uniformly distributed pattern in a highly isotropic manner. The difference with a classical azimuth-elevation sampling can be appreciated in \figref{fig:fibonacci}. Given the \textit{lattice} nature, angles are incrementally generated in the following way:
\begin{align}
\label{eq:fib}
\theta_i = \arccos{\left(\frac{1 - 2i}{N}\right)}, \quad \quad \phi_i = i\pi \left(1 + \sqrt{5} \right), \quad \quad i=0,\dots, N.
\end{align}
We calculate $\bR=\bR_\theta \bR_\phi \in \mathbb{SO}(3)$ making a rotation along the $x$-axis, followed by a rotation along $y$-axis.

\subsection{Visibility Check}
\label{sec:visibility}
After sampling the set of orientations $\mathcal{R}$ for each camera position $\mathbf{t}_j$, it becomes crucial to identify the subset $\cQ_j$ that enables the visibility of 3D landmarks. It is important to discard viewpoints where none or only a minimal number of landmarks are visible.
To accomplish this, we perform an initial visibility check to obtain $\cQ_j \subseteq \cR$ that maximizes visibility. 
Note that $\cQ_j$ is the set calculated for each camera location $j$, hence $\{\cQ_0, \dots, \cQ_M\} \subseteq \cQ $. 

To perform visibility checks, we generate a virtual image for each view $\bT_{ij}$, projecting each landmark in $\cP$, given the camera intrinsics.
A projection is a mapping $\pi : \bbR^3 \rightarrow \Gamma \subset \bbR^2$ from a landmark $\bl$ to image coordinates $\bu = [x_u, y_u]^T$. In our algorithm, we employ the pinhole projection model~\cite{hartley2003multiple}. However, it can be straightforwardly replaced by any other model. 
For each camera location, we keep the best orientations that provide the highest number of visible landmarks. 
In order to detect occlusions, given the sparsity of $\cP$, we perform one of the following steps based on the input (\ie, depending on whether dense depth data is available or not):
\begin{itemize}
    \item with dense depth, we run a \gls{tsdf} integration and store a dense model in a hash table similar to the one originally proposed in \cite{niessner2013real}. 
    This assumes to have generated a dense model other than $\cP$ during mapping. 
    This model can be queried in $O(1)$ given $\bT_{ij}$, and occlusions can be detected using a simple z-buffer algorithm;
    \item without dense depth, we employ a sparse hidden point removal technique \cite{katz2007direct} based only on sparse data $\cP$. Given a viewpoint, this method first transforms the point cloud points to a new range-dependent domain and then constructs the convex hull in that domain.
\end{itemize}
While the second approach does not require additional input, it is expected to be less accurate due to having significantly less information regarding the scene. 

Given the independence of each $\bT_{ij}$, we implemented this initial visibility check on the GPU with CUDA for faster computation. 
After identifying the viewpoints that maximize the visibility of landmarks for each camera position $\bt_j$, we sort the viewpoints in descending order, prioritizing visibility.


\subsection{Training}
\label{sec:nn}
Using the best set
of orientations $\cQ$ we can obtain the full set of viewpoints that maximizes the visibility. For each viewpoint, we compute a feature vector $\bx_k$ that contains the visibility information and an expected value $y_k$ representing if the viewpoint is good enough for localization.
Therefore, our viewpoint evaluation model learns a scoring function $\mathbf{f}_{\cP}(\bR, \bt) \rightarrow [0,1]$, that given a set of input features $\bx$, provides an estimate of the normalized localization quality. 
For learning, we employ a mean weighted binary-cross-entropy loss within a \gls{mlp} encoder, casting the problem to classification:
\begin{equation}
    \cL(\bW) = \frac{1}{|\cD|} \sum_{k=1}^{|\cD|} w_k (y_k\log p_k  + (1 - y_k) \log(1-p_k)),
\end{equation}
where $\cD$ is the dataset $\cD = \{(\bx_k, y_k)\}$. The term $p_k$ denotes each sample's classifier output probability, $y_k$ denotes the correspondence target label, and $w_k$ is the weight balancing the negative and positive samples. Additionally, to address overfitting, we incorporate regularization terms into the loss function:
\begin{equation}
\cL_{\text{total}}(\bW) = \cL(\bW) + \lambda \sum_{l=1}^{L} \| \bW_l \|_2^2,
\end{equation}
where $\bW_l$ represents the weights of layer $l$ in the \gls{mlp}, and $\lambda$ controls the regularization strength.

\subsubsection{Data.}
In order to learn our scoring function $\mathbf{f}_{\cP}(\bR, \bt)$, we provide $\bx_k$ for each camera viewpoint $\bT_k$. This involves only the set of visible landmarks $\in \bbR^{3H}$ and their reprojections in a virtual image $\pi(\bT_{k}^{-1} \cP) \in \bbR^{2H}$. In our experiments, we try different input data, namely $\bx_k
\in \bbR^{2H}$: only landmarks projection, $\bx_k
\in \bbR^{3H}$: image projections and depth data, and $\bx_k \in \bbR^{5H}$: image projections and 3D landmarks in camera frame $(\bT_{k}^{-1} \cP)$. For now, we assume that our best model is the latter; however, in \tabref{tab:ourresults}, we show numerically how this different information impacts localization score. 
A crucial aspect in both localization and image registration is the spatial distribution of features throughout the image \cite{shi1994good}. The more uniform this distribution is across the image, the higher the likelihood of obtaining accurate results in the registration process and the less likely to have a degenerate solution \cite{torr1998robust}. To ensure a consistent and fixed input for our network, particularly given the variations in the number of observable landmarks across diverse viewpoints, we employ image binning (\ie, grouping the reprojected landmarks into discrete cells in the image). 
In our experiments, we set this grid dimension to be $30\times30$ bins. Within each bin, we calculate the mean of visible landmarks both in 3D and in 2D. This technique allows a consistent input that considers the distribution of the features in the image rather than performing, for instance, random sampling. Bins without visible landmarks are systematically assigned a zero value. The proposed strategy allows a uniformly distributed fixed input size for the \gls{mlp} encoder without affecting the complexity of the model and/or additional preprocessing steps.  
Given the heterogenous input (2D projections, 3D landmarks, or depth landmark data) and to speed up the training process, we standardize the input data. For each feature $x$, we calculate its corresponding standardized value $z = \frac{x - \mu}{\sigma}$, where $\mu$ and $\sigma$ are the mean and standard deviation of the kind of input. The benefits of centering the feature values around zero are described in \cite{lecun2002efficient}.


\setlength{\belowcaptionskip}{-20px}
\setlength{\abovecaptionskip}{5px}
\begin{wraptable}{r}{0.50\textwidth}
\vspace{-22px}
\centering
\begin{tabular}{c c} 
  \hline
  Layer & Type \\
  \hline
  1 & fully connected $\text{input-size} \times 300$ \\
  1.1 & ReLU \\
  1.2 & dropout $0.5$ \\
  \hline
  2 & fully connected $300 \times 300$ \\
  2.1 & ReLU \\
  2.2 & dropout $0.5$ \\
  \hline
  3 & fully connected $300 \times 1$ \\
  3.1 & Sigmoid \\
  \hline
\end{tabular}
\caption{\textbf{The architecture} employed to classify the quality of camera viewpoints. }
\label{tab:arch}
\end{wraptable}
\setlength{\belowcaptionskip}{\originalbelowcaptionskip}
\setlength{\abovecaptionskip}{\originalabovecaptionskip}

\subsubsection{Self-supervision.}
In our approach, we leverage our structured framework that involves inputting a \gls{sfm} model and constructing a set of 3D locations denoted as $\cV$ and a collection of orientations represented by $\cQ$ as explained in \secref{sec:sampling}. 
For each viewpoint $\bT_k$ generated through our methodology, we use a simulator relying on Open3D \cite{zhou2018open3d} and Habitat - Matterport 3D meshes \cite{ramakrishnan2021habitat} to extract a dense RGB image. If the depth image can be queried, we use this to filter occlusions for processing the training set; otherwise, we rely on \cite{katz2007direct} (\secref{sec:visibility}).
This acquired image is fundamental to calculating the expected value $y_k$. We achieve this by employing COLMAP \cite{schonberger2016structure} in global localization mode, where the RGB image is localized against the \gls{sfm} model. Subsequently, given the ground-truth pose obtained from the simulator, we assess the error in registering the image against the \gls{sfm} model, considering both rotation and translation. In our setup, 3D landmarks are represented by triangulated SIFT features \cite{lowe2004distinctive}. 
If this error falls below a predefined threshold, we label $y_k$ as a positive sample. Otherwise, it is set to negative. 
This process ensures autonomously the accuracy and reliability of our labeling mechanism, contributing to the overall effectiveness of our methodology.

\section{Experiments}
\label{sec:experiments}
To show the effectiveness of our approach, we conducted different qualitative and quantitative experiments with simulated data and in real-world scenarios. We trained our model using 10 different meshes from Habitat - Matterport 3D \cite{ramakrishnan2021habitat}, and we evaluated the generalization capabilities on 2 different meshes. 
The training set includes around 250k camera viewpoints, balancing negative and positive samples. 
In contrast, the test set, where we evaluate our approach, comprises around 92k viewpoints, which are never shown during training and contain around 70\% of negative samples and 30\% of positive samples.
We train our best model of the size shown in \tabref{tab:arch}, with Adam optimizer and learning rate initialized to 1e-3 for $300$ epochs. 
Taking care of the binning with a grid of $30\times30$, we feed the network with a flattened input containing the projected landmark in 2D, its depth, or its 3D value. 
Hence, the total input length is $4220$. All experiments and training were conducted on a machine equipped with an Intel Core i7-7700K CPU @ 4.20GHz, featuring 8 cores, and a GeForce GTX 1070 GPU.

\subsection{Localization accuracy}

To assess how accurately our method localizes, we use metrics from the Long-Term Visual Localization benchmark \cite{sattler2018benchmarking}. 
Originally designed for outdoor and large-scale settings, the benchmark defined three accuracy ranges (0.25m, 2° / 0.5m, 5° / 5m, 10°). 
Since we focus on indoor scenes, we include an additional finest range (0.05m, 0.4°) to ensure a meaningful evaluation in this environment.  The localization results are reported as the percentage of query images localized within the four given translation and rotation thresholds for each condition.

In our experiments, we compared our method with \gls{fif} \cite{zhang2020fisher} and a few other baseline approaches. 
\gls{fif} utilizes Fisher Information theory, treating the camera as a bearing vector to make information independent from rotation. 
A Gaussian Process regression is then applied to determine a visibility score. As commonly done in information theory \cite{placed2023survey}, \gls{fif} evaluates the information based on the minimum eigenvalue, the determinant, and the trace. 

We report all the results in \tabref{tab:vsothers}.  For baselines, we employed random sampling and reprojection with bins. In the first, we select the target viewpoint randomly, based on the distribution of the test set. In the reprojection with bins approach, we choose the viewpoint that maximizes the number of binned features, ensuring a uniform distribution of features in the image.


Compared to existing methods like \gls{fif}, the main advantage of our approach is that it allows multiple camera viewpoints to be considered ``good'' at each map location. This flexibility means that during planning, one can select the camera viewpoint that offers the most convenience—such as the shortest path—or the best balance between accuracy for localization and planning convenience. In \figref{fig:comparison}, we show how predicting multiple viewpoints for each camera impacts the localization score of the Long-Term Visual Localization benchmark \cite{sattler2018benchmarking}. We specifically experimented with 1, 5, and 10 best directions for each camera location.

\begin{table*}[t]
\centering
\begin{tabular}{rcccc}
\toprule
\toprule
\textbf{Method} & 0.05m, 0.4° / & 0.25m, 2° / & 0.5m, 5° / & 5m, 10° \\
\midrule
random	& 54.7	& 60.9 & 60.9 &	61.1 \\
\gls{fif} mineig exact & 60.1 &	67.7 &	69.0 &	69.6 \\ 
\gls{fif} mineig app	& 57.3 & 	64.7 &	66.2 &	66.4 \\ 
\gls{fif} mineig GP app & 59.8 &	65.4 &	65.7 &	66.8 \\ 
\gls{fif} det exact	& 59.2 & 65.4 & 66.2 & 66.2 \\ 
\gls{fif} det app	& 60.7 & 67.5 & 69.1 & 69.4 \\ 
\gls{fif} det GP app & 58.8 & 67.0 & 67.1 & 67.2 \\ 
\gls{fif} trace app & 55.4 &	61.1 &	61.8 &	62.0 \\ 
\gls{fif} trace GP app & - & - &	- & - \\ 
\gls{fif} trace appr worst & 17.8 &	23.3 &	23.3 &	23.3 \\ 
\gls{fif} trace appr zero deriv labels & 12.5 & 16.7 & 17.3 & 17.3 \\  
\gls{fif} trace exact	& - & - &	- & - \\ 
reprojections with bins &	68.7 &	71.7 &	71.7 &	71.9 \\ 
\textbf{proposed}  	& \textbf{72.9} &	\textbf{80.4} & \textbf{81.2} & \textbf{81.7} \\ 
\bottomrule
\bottomrule
\end{tabular}
\caption{\textbf{Quantitative results.} The localization score values [\%] on a test set of around 92k camera viewpoints, calculated following the Long-Term Visual Localization \cite{sattler2018benchmarking} benchmark with the addition of the finest scale to target specifically our indoor setting. 
We compare against \gls{fif} \cite{zhang2020fisher} using multiple modalities to evaluate the Fisher information matrix, random, and reprojection with bins. The proposed approach leads to the most accurate results. 
The fact that the proposed method, relying on point reprojections grouped into bins, significantly outperforms \gls{fif} demonstrates that the distribution of landmarks is more important than evaluating the Fisher information matrix in visual localization.}
\label{tab:vsothers}
\vspace{-1cm}
\end{table*}

\subsection{Planning experiment}

\begin{figure}[t]
  \centering
  \begin{subfigure}{0.30\linewidth}
    \includegraphics[width=\linewidth]{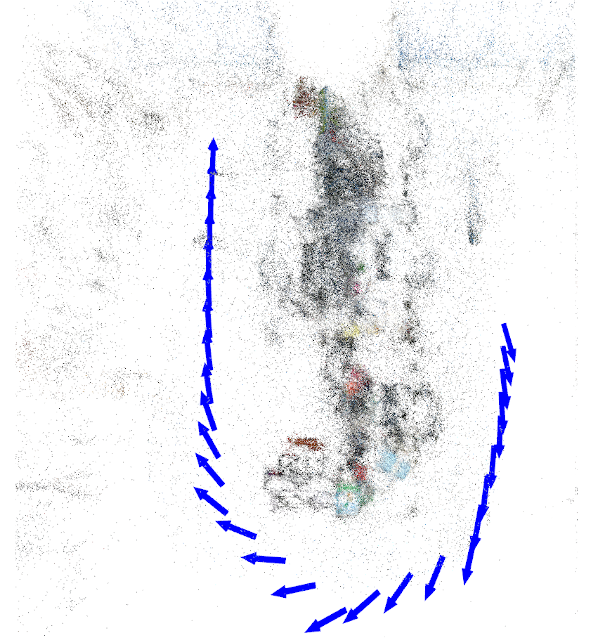}
    \caption{Forward-facing camera}
    \label{fig:sub1}
  \end{subfigure}
  \hfill
  \begin{subfigure}{0.30\linewidth}
    \includegraphics[width=\linewidth]{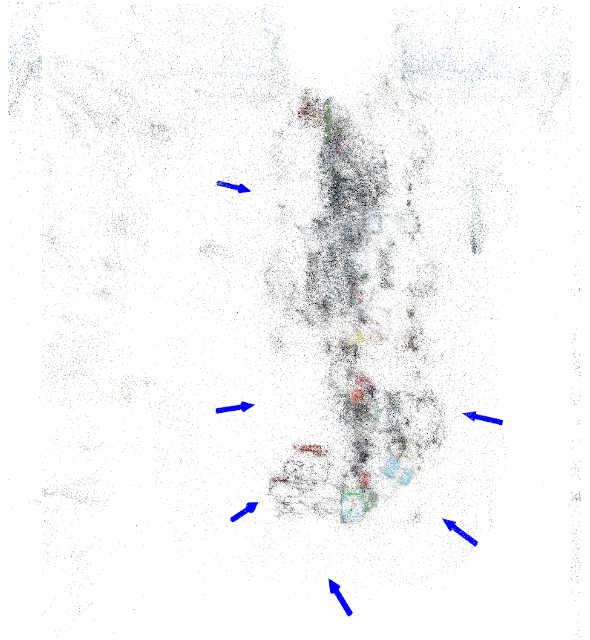}
    \caption{FIF \cite{zhang2020fisher}}
    \label{fig:sub2}
  \end{subfigure}
  \hfill
  \begin{subfigure}{0.30\linewidth}
    \includegraphics[width=\linewidth]{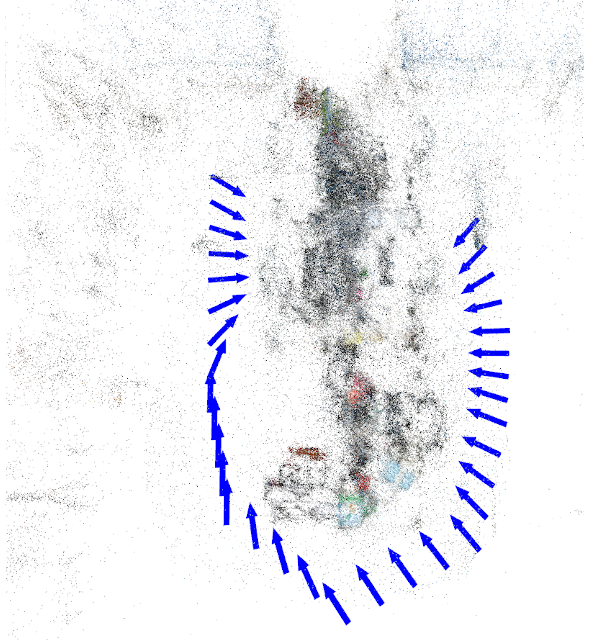}
    \caption{Ours}
    \label{fig:sub3}
  \end{subfigure}
  \caption{\textbf{Qualitative planning} experiments with self-recorded data. In this setup, \gls{fif} encounters challenges in predicting certain viewpoints, leading to failures in our planner. The conventional camera look-forward approach, a traditional method, neglects the consideration of active viewpoints. Its limitations become apparent as it neglects camera adjustments toward regions with higher landmark density, potentially directing it toward featureless areas. We specifically use a noisy \gls{sfm} model to show the robustness and adaptability of our approach, reflecting real case experiments.}
  \label{fig:planning}
  \vspace{-0.4cm}
\end{figure}

We assess the planning quality qualitatively. For 3D planning, we utilized RRT* \cite{karaman2011sampling}, a planner that iteratively grows the tree by sampling random configurations in the configuration space, with our state represented as $\mathbb{SE}(3)$. In our planning comparisons, we used Fisher information fields and a classic camera look-forward approach. During experiments, \gls{fif} fails to predict some viewpoints, resulting in a failure state. The conventional camera look-forward approach is a traditional method that overlooks the consideration of active viewpoints, and its drawbacks become evident as it fails to adjust the camera towards regions with higher landmark density, potentially pointing towards featureless areas. In contrast, our approach achieves meaningful planning, directing the camera towards regions with higher landmark density. These qualitative results are depicted in \figref{fig:planning}. We specifically use a noisy \gls{sfm} model to show the robustness and adaptability of our approach, reflecting real case experiments. The \gls{sfm} model has been generated using only sparse COLMAP \cite{schonberger2016structure} reconstruction, without given poses, using only a set of images as input. We used markers to retrieve the original scale.

\setlength{\belowcaptionskip}{-20px}
\setlength{\abovecaptionskip}{-5px}
\begin{wrapfigure}{r}{0.40\textwidth}
\vspace{-60px}
\begin{center}
\includegraphics[width=0.40\textwidth]{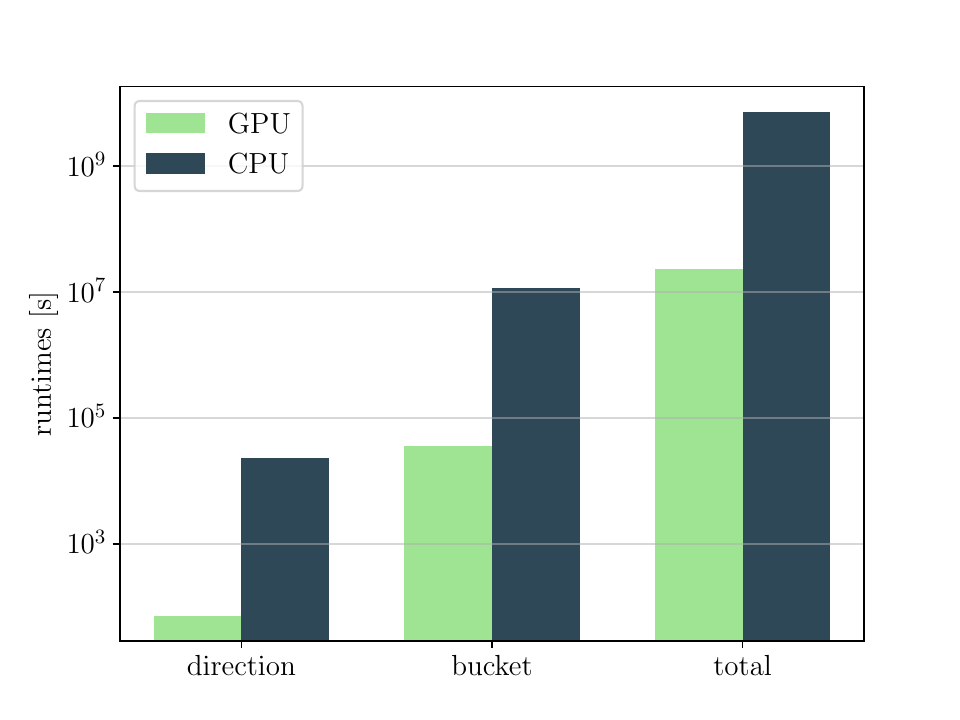}
\end{center}
\caption{\textbf{Runtimes} of our GPU sampling method compared with a single-threaded CPU implementation. These experiments are done by sampling from 648 camera locations with 24919 3D landmarks from the \gls{sfm} model.}
\label{fig:runtimes}
\end{wrapfigure}
\setlength{\belowcaptionskip}{\originalbelowcaptionskip}
\setlength{\abovecaptionskip}{\originalabovecaptionskip}

\subsection{Runtimes}
We adopt a simple model to learn the active camera viewpoints to make inferences promptly, designed for dedicated devices and robotics. Although a complete map representation can be pre-generated, during motion planning or localization, one can simply query the hash table based on the map location. One inference step of our model takes around 0.02 seconds with our setup. In addition, given the exhaustive approach adopted to subsample good camera viewpoints described in \secref{sec:space-sampling}, and given the independence of each camera viewpoint, we make our implementation in CUDA leveraging parallelism. 

In \figref{fig:runtimes}, we illustrate the runtimes of our sampling compared to a single-threaded CPU implementation, presenting the plot in $\log$ scale for clarity. Numerically, the GPU implementation completes the initial sampling in around 23 seconds, while the CPU implementation takes approximately 7387 seconds. These experiments involve sampling from 648 camera locations with 24919 landmarks from the \gls{sfm} model. 


\section{Ablation Study}

Within this study, our primary focus is on extracting valuable insights from environmental geometry to advance active visual localization. For this, we introduce a data-driven approach employing a compact architecture designed for real-time operations, a novel self-supervised training method, the development of a unique map representation allowing specific voxel locations in space to possess one or more active viewpoints, and the possibility to integrate motion planning for robotics applications.

\begin{figure}[t]
  \centering
  \includegraphics[width=\textwidth]{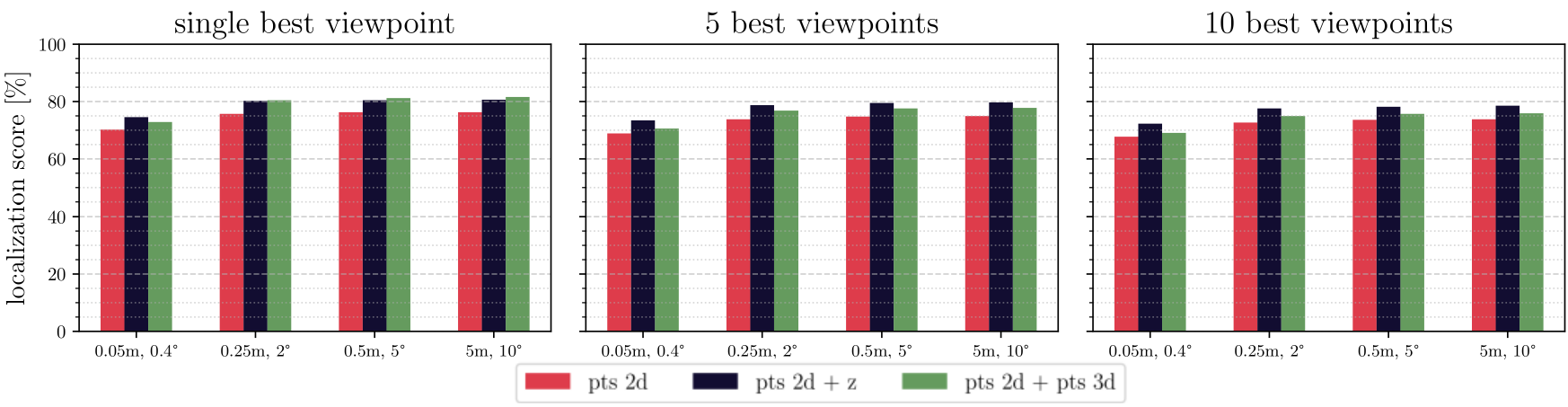}
  \caption{\textbf{Multiple viewpoints results}.  
  This figure demonstrates the impact of predicting multiple viewpoints on localization score according to \cite{sattler2018benchmarking}, experimenting with 1, 5, and 10 directions per location. Compared to methods like \gls{fif}, our approach's main advantage is allowing multiple ``good'' camera viewpoints at each map location. 
  This flexibility lets planners choose the most convenient viewpoint, such as the shortest path or the best balance between localization accuracy and convenience.
  In this experiment, we collected all the predicting viewpoints for each camera location, sorted the likelihood output from the \gls{mlp} encoder in descending order, and took the best $N$ elements for evaluation.}
  \label{fig:comparison}
\end{figure}

The discrete map representation presented in this work using a voxel grid, parameterizing each camera location as a voxel, is similar to the one proposed in \cite{zhang2018perception}. The main difference is that within our active viewpoint selection, the independence of each viewpoint in the classification process allows the possibility of selecting more camera viewpoints for each voxel location. 
This makes things easier during motion planning, for example, because the planner can rely on the viewpoint in the location that minimizes the cost concerning its current state (\ie~shortest path). How predicting multiple directions within our methodology from each camera location impacts the localization score is analyzed in \figref{fig:comparison}.

The results obtained in our experiments demonstrate how the distribution of observed landmarks in the image impacts visual localization tasks. 
This seems more important than exploiting the Fisher information matrix, which usually gives more importance to the vicinity, generally maximizing visibility, without considering the distribution of the observed map in the image. 

In our experiments, we investigated how incorporating various types of information affects the quality of viewpoint selection. We explored scenarios with only 2D visible landmarks and introduced the full 3D landmarks (in relative camera frame) or only their depth. The detailed results are presented in \tabref{tab:ourresults}. Notably, 3D information significantly enhances our model's accuracy. In general, depth information alone is sufficient to achieve good results. While incorporating the full 3D geometry can improve performance, it may also lead to overfitting, likely due to data quality issues or its limited usefulness for generalization. 

We evaluated our model through quantitative testing on simulated data (\tabref{tab:vsothers}) and qualitative analysis on real data (\figref{fig:planning}). It is worth noting that our approach never explicitly generates a complete RGB image during training (it is used just to calculate the expected value for supervision). Instead, it relies solely on geometric information, making it independent of specific digital details encoded in RGB images. This characteristic significantly enhances the model's generalization ability, enabling transitions from photorealistic simulated data to real-world scenarios. This capability is demonstrated in the planning experiments conducted with real data illustrated in \figref{fig:planning}.

We specifically use a simple MLP encoder since it does not require any preprocessing, unlike, for example, a \gls{gnn}, where a graph for each sample must be created, or a \gls{cnn}, where the full image should be convoluted. Also, training and inference are fast. While more complex models might offer better accuracy, exploiting the correlation between landmarks, we aim to keep the approach simple to enhance its adaptability to real conditions.

\begin{table*}[h]
\centering
\begin{tabular}{ccccc}
\toprule
\toprule
\textbf{approach} & 0.05m, 0.4° / & 0.25m, 2°/ & 0.5m, 5° / & 5m, 10° \\
\midrule
ours  (pts 2d) & 70.2 & 75.7 & 76.2 & 76.2 \\ 
ours (pts 2d + z) & \textbf{74.6} & 80.3 & 80.4 & 80.6 \\ 
ours  (pts 2d + pts 3d) & 72.9 & \textbf{80.4} & \textbf{81.2} & \textbf{81.7} \\ 
\bottomrule
\bottomrule
\end{tabular}
\caption{\textbf{Impact of different geometrical information} in our viewpoint selection strategy. We examined situations first with only 2D visible landmarks, then introduced landmark depth (along \textit{z}-camera axis), and finally, considered the full 3D points. Intuitively, the 3D data improves localization scores compared to 2D-only cases. Localization score according to \cite{sattler2018benchmarking} is reported in \%.}
\label{tab:ourresults}
\end{table*}

\vspace{-1.7cm}


\section{Conclusion}

\label{sec:conclusion}
This paper explores active localization, highlighting viewpoint selection's crucial role in refining localization accuracy. Our contributions involve a data-driven approach with a simple architecture designed for real-time operation, introducing a self-supervised data training method. We show the capabilities of our viewpoints map to be integrated into a planning framework for robotics applications. We conducted both qualitative and numerical experiments on simulated and real data. Our results demonstrate the performance of our method compared to existing solutions for similar challenges, proving its effectiveness. For the future, we envision a more robust model, developed ad-hoc for selecting the best camera viewpoints and an active localization benchmark to benefit the community.

\section*{Acknowledgments}
This work has been partially supported by Sapienza University of Rome as part of the work for project \textit{H\&M: Hyperspectral and Multispectral Fruit Sugar Content Estimation for Robot Harvesting Operations in Difficult Environments}, Del. SA n.36/2022, by the Hasler Stiftung Research Grant via the ETH Zurich Foundation and an ETH Zurich Career Seed Award.





%
%
\bibliographystyle{splncs04}
\bibliography{egbib}
\end{document}

%% file: notation.tex
\usepackage{amsopn}

\newcommand{\bl}{\mathbf{l}}
\newcommand{\bt}{\mathbf{t}}

\newcommand{\bW}{\mathbf{W}}

\newcommand{\bR}{\mathbf{R}}

\newcommand{\bT}{\mathbf{T}}

\newcommand{\cD}{\mathcal{D}}

\newcommand{\cR}{\mathcal{R}}

\newcommand{\cV}{\mathcal{V}}

\newcommand{\cP}{\mathcal{P}}
\newcommand{\cL}{\mathcal{L}}

\newcommand{\cQ}{\mathcal{Q}}

\newcommand{\bbR}{\mathbb{R}}

\newcommand{\bbSO}{\mathbb{SO}}

\newcommand{\bx}{\mathbf{x}}

\newcommand{\bu}{\mathbf{u}}

\newcommand{\bzero}{\mathbf{0}}

\def\g2o{$g^2o$}
\def\t2v{\mathrm{log}}
\def\v2t{\mathrm{exp}}
\def\ev2t{\mathrm{ev2t}}

\def\ie{{i.e.}}
\def\eg{{e.g.}}

\def\secref#1{Sec.~\ref{#1}}
\def\figref#1{Fig.~\ref{#1}}
\def\tabref#1{Tab.~\ref{#1}}
\def\eqref#1{Eq.~(\ref{#1})}

